\documentstyle[aclap,psfig]{article}
\title{A Portable Algorithm for Mapping Bitext Correspondence}
\author{I. Dan Melamed \\ Dept. of Computer and Information Science \\
University of Pennsylvania \\ Philadelphia, PA, 19104, U.S.A. \\
{\tt melamed@unagi.cis.upenn.edu}}

\date{}
\begin{document}
\maketitle

\begin{abstract}
The first step in most empirical work in multilingual NLP is to
construct maps of the correspondence between texts and their
translations ({\bf bitext maps}).  The Smooth Injective Map Recognizer
(SIMR) algorithm presented here is a generic pattern recognition
algorithm that is particularly well-suited to mapping bitext
correspondence.  SIMR is faster and significantly more accurate than
other algorithms in the literature.  The algorithm is robust enough to
use on noisy texts, such as those resulting from OCR input, and on
translations that are not very literal.  SIMR encapsulates its
language-specific heuristics, so that it can be ported to any language
pair with a minimal effort.
\end{abstract}

\section{Introduction}

Texts that are available in two languages (bitexts) are immensely
valuable for many natural language processing applications\footnote{
``Multitexts'' in more than two languages are even more valuable, but
they are much more rare.}.  Bitexts are the raw material from which
translation models are built.  In addition to their use in machine
translation \cite{ebmt,ibm,transmod}, translation models can be
applied to machine-assisted translation \cite{sato,transtype},
cross-lingual information retrieval \cite{clir}, and gisting of World
Wide Web pages \cite{gisting}.  Bitexts also play a role in less
automated applications such as concordancing for bilingual
lexicography \cite{catiz,wordcorr}, computer-assisted language
learning, and tools for translators (e.g. \cite{mack95,adomit}.
However, bitexts are of little use without an automatic method for
constructing bitext maps.

Bitext maps identify corresponding text units between the two halves
of a bitext.  The ideal bitext mapping algorithm should be fast and
accurate, use little memory and degrade gracefully when faced with
translation irregularities like omissions and inversions.  It should
be applicable to any text genre in any pair of languages.  

The Smooth Injective Map Recognizer (SIMR) algorithm presented in this
paper is a bitext mapping algorithm that advances the state of the art
on these criteria.  The evaluation in Section~\ref{eval} shows that
SIMR's error rates are lower than those of other bitext mapping
algorithms by an order of magnitude.  At the same time, its expected running
time and memory requirements are linear in the size of the input,
better than any other published algorithm.

The paper begins by laying down SIMR's geometric foundations and
describing the algorithm.  Then, Section~\ref{port} explains how to port
SIMR to arbitrary language pairs with minimal effort, without relying
on genre-specific information such as sentence boundaries.  The last
section offers some insights about the optimal level of text analysis
for mapping bitext correspondence.

\section{Bitext Geometry}
\label{form}

A {\bf bitext} \cite{bitext} comprises two versions of a text, such as
a text in two different languages.  Translators create a bitext each
time they translate a text.
\begin{figure}[htb]
\centerline{\psfig{figure=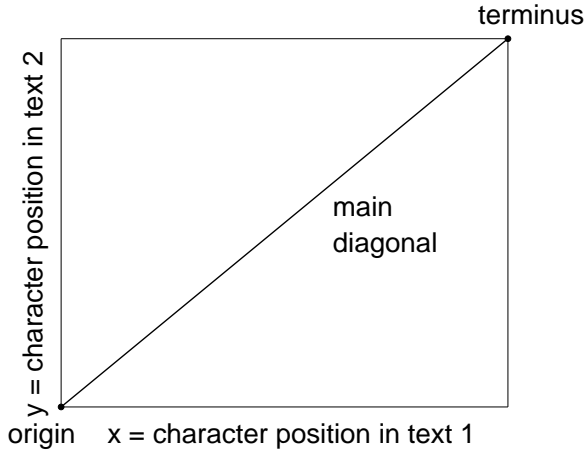,width=3in}}
\caption{{\em a bitext space }}
\label{corresp}
\end{figure}
Each bitext defines a rectangular {\bf bitext~space}, as
illustrated in Figure~\ref{corresp}.  The width and height of the
rectangle are the lengths of the two component texts, in characters.
The lower left corner of the rectangle is the {\bf origin} of the
bitext space and represents the two texts' beginnings.  The upper
right corner is the {\bf terminus} and represents the texts' ends.
The line between the origin and the terminus is the {\bf main
diagonal}.  The slope of the main diagonal is the {\bf bitext slope}.

Each bitext space contains a number of {\bf true points of
correspondence (TPCs)}, other than the origin and the terminus.  For
example, if a token at position $p$ on the x-axis and a token at
position $q$ on the y-axis are translations of each other, then the
coordinate $(p, q)$ in the bitext space is a TPC\footnote{Since
distances in the bitext space are measured in characters, the position
of a token is defined as the mean position of its characters.}.
TPCs also exist at corresponding boundaries of text units such as
sentences, paragraphs, and chapters.  Groups of TPCs with a roughly
linear arrangement in the bitext space are called {\bf chains}.

{\bf Bitext maps} are 1-to-1 functions in bitext spaces.  A
complete set of TPCs for a particular bitext is called a {\bf true
bitext map (TBM)}.  The purpose of a {\bf bitext mapping algorithm} is
to produce bitext maps that are the best possible approximations of
each bitext's TBM.

\section{SIMR}
\label{simr}
SIMR builds bitext maps one chain at a time.  The search for each
chain alternates between a generation phase and a recognition phase.
The generation phase begins in a small rectangular region of the
bitext space, whose diagonal is parallel to the main diagonal.  Within
this search rectangle, SIMR generates all the points of correspondence
that satisfy the supplied matching predicate, as explained in
Section~\ref{matchpred}.  In the recognition phase, SIMR calls the
chain recognition heuristic to find suitable chains among the
generated points.  If no suitable chains are found, the search
rectangle is proportionally expanded and the generation-recognition
cycle is repeated.  The rectangle keeps expanding until at least one
acceptable chain is found. If more than one chain is found in the same
cycle, SIMR accepts the one whose points are least dispersed around
its least-squares line.  Each time SIMR accepts a chain, it selects
another region of the bitext space to search for the next chain.

\begin{figure}[htb]
\centerline{\psfig{figure=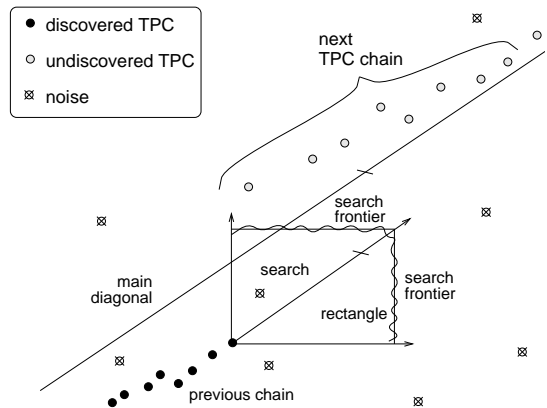,width=3in}}
\caption{{\em SIMR's ``expanding rectangle'' search strategy.  The
search rectangle is anchored at the top right corner of the previously
found chain.  Its diagonal remains parallel to the main diagonal.}}
\label{exprect}
\end{figure}

SIMR employs a simple heuristic to select regions of the bitext space
to search.  To a first approximation, TBMs are monotonically
increasing functions.  This means that if SIMR finds one chain,
it should look for others either above and to the right or below and
to the left of the one it has just found.  All SIMR needs is a place
to start the trace.  A good place to start is at the beginning:
Since the origin of the bitext space is always a TPC, the first search
rectangle is anchored at the origin.  Subsequent search rectangles are
anchored at the top right corner of the previously found chain, as
shown in Figure~\ref{exprect}.

The expanding-rectangle search strategy makes SIMR robust
in the face of TBM dis\-con\-tinui\-ties.  Figure~\ref{exprect} shows
a segment of the TBM that contains a vertical gap (an omission in the
text on the x-axis).  As the search rectangle grows, it will
eventu\-ally intersect with the TBM, even if the dis\-con\-tinui\-ty
is quite large \cite{adomit}.  The noise filter described in
Section~\ref{nf} prevents SIMR from being led astray by false points
of correspondence.

\subsection{Point Generation}
\label{matchpred}
SIMR generates candidate points of correspondence in the search
rectangle using one of its matching predicates.  A {\bf matching
predicate} is a heuristic for deciding whether a given pair of tokens
are likely to be mutual translations.  Two kinds of information that a
matching predicate can rely on most often are cognates and translation
lexicons.

Two tokens in a bitext are {\bf cognates} if they have the same
meaning and similar spellings. In the non-technical Canadian Hansards
(parliamentary debate transcripts available in English and in French),
cognates can be found for roughly one quarter of all text tokens
\cite{melamed95}.  Even distantly related languages like English and
Czech will share a large number of cognates in the form of proper
nouns.  Cognates are more common in bitexts from more similar language
pairs, and from text genres where more word borrowing occurs, such as
technical texts.  When dealing with language pairs that have
dissimilar alphabets, the matching predicate can employ phonetic
cognates \cite{techrep}.  When one or both of the languages involved
is written in pictographs, cognates can still be found among
punctuation and digit strings.  However, cognates of this last kind
are usually too sparse to suffice by themselves.

When the matching predicate cannot generate enough candidate
correspondence points based on cognates, its signal can be
strengthened by a translation lexicon.  Translation lexicons can be
extracted from machine-readable bilingual dictionaries (MRBDs), in the
rare cases where MRBDs are available.  In other cases, they can be
constructed automatically or semi-automatically using any of several
methods \cite{funa,amta,anlp97}.  Since the matching
predicate need not be perfectly accurate, the translation lexicons need not be
either.

Matching predicates can take advantage of other information, besides
cognates and translation lexicons.  For example, a list of {\em faux
amis} is a useful complement to a cognate matching strategy
\cite{mack95}.  A stop list of function words is also helpful.
Function words are translated inconsistently and make unreliable
points of correspondence \cite{techrep}.

\subsection{Point Selection}
\label{select}
As illustrated in Figure~\ref{exprect}, even short sequences of TPCs
form characteristic patterns.  Most chains of TPCs have the following
properties:
\begin{itemize}
\item {\bf Linearity}: TPCs tend to line up straight.  
\item {\bf Low Variance of Slope}: The slope of a TPC chain is rarely
much different from the bitext slope.
\item {\bf Injectivity:} No two points in a chain of TPCs can have the
same x-- or y--co-ordinates.
\end{itemize}
SIMR's chain recognition heuristic exploits these properties to decide
which chains in the search rectangle might be TPC chains.  

The heuristic involves three parameters: {\bf chain size}, {\bf
maximum point dispersal} and {\bf maximum angle deviation}.  A chain's
size is simply the number of points it contains.  The heuristic
considers only chains of exactly the specified size whose points are
injective.  The linearity of the these chains is tested by measuring
the root mean squared distance of the chain's points from the chain's
least-squares line.  If this distance exceeds the maximum point
dispersal threshold, the chain is rejected.  Next, the angle of each
chain's least-squares line is compared to the arctangent of the bitext
slope.  If the difference exceeds the maximum angle deviation
threshold, the chain is rejected.  These filters can be efficiently
combined so that SIMR's expected running time and memory requirements
are linear in the size of the input bitext \cite{techrep}.

The chain recognition heuristic pays no attention to whether chains
are monotonic.  Non-monotonic TPC chains are quite common, because
even languages with similar syntax like French and English have
well-known differences in word order.  For example, English
(adjective, noun) pairs usually correspond to French (noun, adjective)
pairs.  Such inversions result in TPCs arranged like the middle two
points in the ``previous chain'' of Figure~\ref{exprect}.  SIMR has no
problem accepting the inverted points.

If the order of words in a certain text passage is radically altered
during translation, SIMR will simply ignore the words that ``move too
much'' and construct chains out of those that remain more stationary.
The maximum point dispersal parameter limits the width of accepted
chains, but nothing limits their length.  In practice, the chain
recognition heuristic often accepts chains that span several
sentences.  The ability to analyze non-monotonic points of
correspondence over variable-size areas of bitext space makes SIMR
robust enough to use on translations that are not very literal.

\subsection{Noise Filter}
\label{nf}
\begin{figure}[tb]
\centerline{\psfig{figure=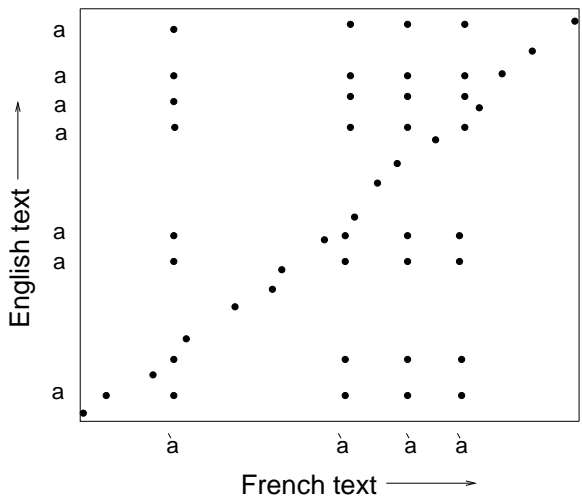,width=3in}}
\caption{{\em Frequent tokens cause false points of
correspondence that line up in rows and columns.}}
\label{noise}
\end{figure}
Points of correspondence among frequent token types often line up in
rows and columns, as illustrated in Figure~\ref{noise}.  Token types
like the English article ``a'' can produce one or more correspondence
points for almost every sentence in the opposite text.  Only one point
of correspondence in each row and column can be correct; the rest are
noise.  A noise filter can make it easier for SIMR to find TPC chains.

Other bitext mapping algorithms mitigate this source of noise either
by assigning lower weights to correspondence points associated with
frequent token types \cite{charalign} or by deleting frequent token
types from the bitext altogether \cite{dagan}.  However, a token type
that is relatively frequent overall can be rare in some parts of the
text.  In those parts, the token type can provide valuable clues to
correspondence.  On the other hand, many tokens of a relatively rare
type can be concentrated in a short segment of the text, resulting in
many false correspondence points.  The varying concentration of
identical tokens suggests that more localized noise filters would be
more effective.  SIMR's localized search strategy provides a vehicle
for a localized noise filter.

The filter is based on the {\bf maximum point ambiguity level}
parameter.  For each point $p = (x, y)$, let X be the number of points
in column $x$ within the search rectangle, and let Y be the number of
points in row $y$ within the search rectangle.  Then the ambiguity
level of $p$ is X + Y -- 2.  In particular, if $p$ is the only point
in its row and column, then its ambiguity level is zero.  The chain
recognition heuristic ignores points whose ambiguity level is too
high.  What makes this a localized filter is that only points within
the search rectangle count toward each other's ambiguity level.  The
ambiguity level of a given point can change when the search rectangle
expands or moves.

\begin{figure}[htb]
\centerline{\psfig{figure=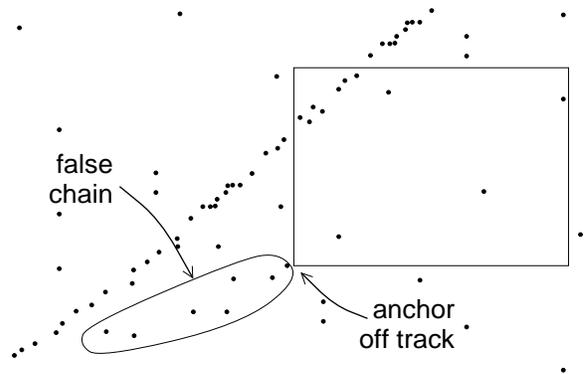,width=3in}}
\caption{{\em SIMR's noise filter ensures that TPCs are much more
dense than false points of correspondence.  A good signal-to-noise
ratio prevents SIMR from getting lost.}}
\label{offtrack}
\end{figure}
The noise filter ensures that false points of correspondence are very
sparse, as illustrated in Figure~\ref{offtrack}.  Even if one chain of
false points of correspondence slips by the chain recognition
heuristic, the expanding rectangle will find its way back to the TBM
before the chain recognition heuristic accepts another chain.  If the
matching predicate generates a reasonably strong signal then the
signal-to-noise ratio will be high and SIMR will not get lost, even
though it is a greedy algorithm with no ability to look ahead.

\section{Porting to New Language Pairs}
\label{port}

SIMR can be ported to a new language pair in three steps.

\subsection{Step 1:  Construct Matching Predicate}
The original SIMR implementation for \linebreak French/English
included matching predicates that could use cognates and/or
translation lexicons.  For language pairs in which lexical cognates
are frequent, a cognate-based matching predicate should suffice.  In
other cases, a ``seed'' translation lexicon may be used to boost the
number of candidate points produced in the generation phase of the
search.  The SIMR implementation for Spanish/English uses only
cognates.  For Korean/English, SIMR takes advantage of punctuation and
number cognates but supplements them with a small translation lexicon.

\subsection{Step 2:  Construct Axis Generators}
In order for SIMR to generate candidate points of correspondence, it
needs to know what token pairs correspond to co-ordinates in the
search rectangle.  It is the axis generator's job to map the two
halves of the bitext to positions on the x- and y-axes of the bitext
space, before SIMR starts searching for chains.  This mapping should
be done with the matching predicate in mind.

If the matching predicate uses cognates, then every word that might
have a cognate in the other half of the bitext should be assigned its
own axis position.  This rule applies to punctuation and numbers as
well as to ``lexical'' cognates.  In the case of lexical cognates, the
axis generator typically needs to invoke a language-specific
tokenization program to identify words in the text.  Writing such a
program may constitute a significant part of the porting effort, if no
such program is available in advance.  The effort may be lessened,
however, by the realization that it is acceptable for the tokenization
program to overgenerate just as it is acceptable for the matching predicate.
For example, when tokenizing German text, it is not necessary for the
tokenizer to know which words are compounds.  A word that has another
word as a substring should result in one axis position for
the substring and one for the superstring.

When lexical cognates are not being used, the axis generator only
needs to identify punctuation, numbers, and those character strings in
the text which also appear on the relevant side of the translation
lexicon\footnote{Multi-word expressions in the translation lexicon are
treated just like any other character string.}.  It would be pointless
to plot other words on the axes because the matching predicate could
never match them anyway.  Therefore, for languages like Chinese and
Japanese, which are written without spaces between words, tokenization
boils down to string matching.  In this manner, SIMR circumvents the
difficult problem of word identification in these languages.

\subsection{Step 3:  Re-optimize Parameters}

The last step in the porting process is to re-optimize SIMR's
numerical parameters.  The four parameters described in
Section~\ref{simr} interact in complicated ways, and it is impossible
to find a good parameter set analytically.  It is easier to optimize
these parameters empirically, using simulated annealing \cite{simann}.

Simulated annealing requires an objective function to optimize.  The
objective function for bitext mapping should measure the difference
between the TBM and maps produced with the current parameter set.  In
geometric terms, the difference is a distance.  The TBM consists of a
set of TPCs.  The error between a bitext map and each TPC can be
defined as the horizontal distance, the vertical distance, or the
distance perpendicular to the main diagonal.  The first two
alternatives would minimize the error with respect to only one
language or the other.  The perpendicular distance is a more robust
average.  In order to penalize large errors more heavily, root mean
squared (RMS) distance is minimized instead of mean distance.

The most tedious part of the porting process is the construction of
TBMs against which SIMR's parameters can be optimized and tested.  The
easiest way to construct these gold standards is to extract them from
pairs of hand-aligned text segments: The final character positions of
each segment in an aligned pair are the co-ordinates of a TPC.  Over
the course of two porting efforts, I have developed and refined tools
and methods that allow a bilingual annotator to construct the required
TBMs very efficiently from a raw bitext.  For example, a tool
originally designed for automatic detection of omissions in
translations \cite{adomit} was adopted to detect mis-alignments.

\subsection{Porting Experience Summary}

Table~\ref{labor} summarizes the amount of time invested in each
new language pair.
\begin{table*}[htb]
\centering
\caption{{\em Time spent in constructing two ``gold standard'' TBMs.}}
\label{labor}
\begin{tabular}{|c|c|c|c|c|}
\hline
& & estimated time & estimated time & number of \\
& main informant for & spent to build & spent on & segments \\
language pair & matching predicate & new axis generator & hand-alignment & aligned \\ \hline \hline
Spanish/English & lexical cognates & 8 h & 5 h & 1338 \\
Korean/English & translation lexicon & 6 h & 12 h & 1224 \\
\hline
\end{tabular}
\end{table*}
The estimated times for building axis generators do not include the
time spent to build the English axis generator, which was part of the
original implementation.  Axis generators need to be built only once
per language, rather than once per language pair.

\section{Evaluation}
\label{eval}

\begin{table*}[htb]
\centering
\caption{{\em SIMR accuracy on different text genres in three language
pairs.}}
\label{results}
\begin{tabular}{|c|c|c|c|c|}
\hline
language & number of & & number of & RMS Error \\
pair & training TPCs & genre & test TPCs & in characters \\ \hline \hline
French / English & 598 & parliamentary debates & 7123 & 5.7 \\
&  & CITI technical reports & 365, 305, 176 & 4.4, 2.6, 9.9 \\
&   & other technical reports & 561, 1393 & 20.6, 14.2 \\
&   & court transcripts & 1377 & 3.9 \\
&   & U.N. annual report & 2049 & 12.36 \\
&   & I.L.O. report & 7129 & 6.42 \\ \hline
Spanish / English & 562 & software manuals & 376, 151, 100, 349 & 4.7,
1.3, 6.6, 4.9 \\ \hline
Korean / English & 615 & military manuals  & 40, 88, 186, 299 & 2.6, 7.1, 25, 7.8\\
&  & military messages & 192 & 0.53 \\
\hline
\end{tabular}
\end{table*}
SIMR was evaluated on hand-aligned bitexts of various genres in three
language pairs.  None of these test bitexts were used anywhere in the
training or porting procedures.  Each test bitext was converted to a
set of TPCs by noting the pair of character positions at the end of
each aligned pair of text segments.  The test metric was the root mean
squared distance, in characters, between each TPC and the interpolated
bitext map produced by SIMR, where the distance was measured
perpendicular to the main diagonal.  The results are presented in
Table~\ref{results}.

\begin{table}
\centering
\caption{{\em SIMR's error distribution on the French/English ``parliamentary
debates'' bitext.} \label{distrib}}
\begin{tabular}{|r|c|c|}
\hline
number of & error range & fraction of \\
test points  & in characters & test points \\ \hline
1       & -101   & .0001 \\
2       & -80 to -70     & .0003 \\
1       & -70 to -60     & .0001 \\
5       & -60 to -50     & .0007 \\
4       & -50 to -40     & .0006 \\
6       & -40 to -30     & .0008 \\
9       & -30 to -20     & .0013 \\
29      & -20 to -10     & .0041 \\
3057    & -10 to 0       & .4292 \\
3902    & 0 to 10        & .5478 \\
43      & 10 to 20       & .0060 \\
28      & 20 to 30       & .0039 \\
17      & 30 to 40       & .0024 \\
5       & 40 to 50       & .0007 \\
8       & 50 to 60       & .0011 \\
1       & 60 to 70       & .0001 \\
1       & 70 to 80       & .0001 \\
1       & 80 to 90       & .0001 \\
1       & 90 to 100      & .0001 \\
1       & 110 to 120     & .0001 \\
1       & 185     & .0001 \\
\hline
7123 & & 1.000 \\ \hline
\end{tabular}
\end{table}
The French/English part of the evaluation was performed on bitexts
from the publicly available BAF corpus created at CITI \cite{BAF}.
SIMR's error distribution on the ``parliamentary debates'' bitext in
this collection is given in Table~\ref{distrib}.  This distribution
can be compared to error distributions reported in \cite{charalign}
and in \cite{dagan}.  SIMR's RMS error on this bitext was 5.7
characters.  Church's {\tt char\_align} algorithm \cite{charalign} is
the only algorithm that does not use sentence boundary information for
which comparable results have been reported.  {\tt char\_align}'s RMS
error on this bitext was 57 characters, exactly ten times higher.

Two teams of researchers have reported results on the same
``parliamentary debates'' bitext for algorithms that map
correspondence at the sentence level \cite{align,simard}.  Both of
these algorithms use sentence boundary information.  Melamed (1996a)
showed that sentence boundary information can be used to convert
SIMR's output into sentence alignments that are more accurate than
those obtained by either of the other two approaches.

The test bitexts in the other two language pairs were created when 
SIMR was being ported to those languages.  The Spanish/English bitexts
were drawn from the on-line Sun MicroSystems Solaris AnswerBooks.  The
Korean/English bitexts were provided and hand-aligned by Young-Suk Lee
of MIT's Lincoln Laboratories.  Although it is not possible to compare
SIMR's performance on these language pairs to the performance of other
algorithms, Table~\ref{results} shows that the performance on other
language pairs is no worse than performance on French/English.

\section{Which Text Units to Map?}

Early bitext mapping algorithms focused on sentences
\cite{tta,debili}.  Although sentence maps do not have sufficient
resolution for some important bitext applications
\cite{adomit,mack95}, sentences were an easy starting point, because
their order rarely changes during translation.  Therefore, sentence
mapping algorithms need not worry about crossing correspondences.  In
1991, two teams of researchers independently discovered that sentences
can be accurately aligned by matching sequences with similar lengths
\cite{align,ibmalign}.

Soon thereafter, Church (1993) found that bitext mapping
at the sentence level is not an option for noisy bitexts found in the
real world.  Sentences are often difficult to detect, especially where
punctuation is missing due to OCR errors.  More importantly, bitexts
often contain lists, tables, titles, footnotes, citations and/or
mark-up codes that foil sentence alignment methods.  Church's solution
was to look at the smallest of text units --- characters --- and to
use digital signal processing techniques to grapple with the much
larger number of text units that might match between the two halves of
a bitext.  Characters match across languages only to the extent that
they participate in cognates.  Thus, Church's method is only
applicable to language pairs with similar alphabets.

The main insight of the present work is that words are a happy
medium-sized text unit at which to map bitext correspondence.  By
situating word positions in a bitext space, the geometric heuristics
of sentence alignment algorithms can be exploited equally well at the
word level.  The cognate heuristic of the character-based algorithms
works better at the word level, because cognateness can be defined
more precisely in terms of words, e.g. using the Longest Common
Subsequence Ratio \cite{melamed95}.  Several other matching heuristics
can only be applied at the word level, including the localized noise
filter in Section~\ref{nf}, lists of stop words and lists of {\em faux
amis} \cite{mack95}.  Most importantly, translation lexicons can only
be used at the word level.  SIMR can employ a small hand-constructed
translation lexicon to map bitexts in any pair of languages, even when
the cognate heuristic is not applicable and sentences cannot be found.
The particular combination of heuristics described in
Section~\ref{simr} can certainly be improved on, but research
into better bitext mapping algorithms is likely to be most fruitfull
at the word level.

\section{Conclusion}
The Smooth Injective Map Recognizer (SIMR) bitext mapping algorithm
advances the state of the art on several frontiers.  It is
significantly more accurate than other algorithms in the literature.
Its expected running time and memory requirements are linear in the
size of the input, which makes it the algorithm of choice for very
large bitexts.  It is not fazed by word order differences.  It does
not rely on pre-segmented input and is portable to any pair of
languages with a minimal effort.  These features make SIMR the mostly
widely applicable bitext mapping algorithm to date.

SIMR opens up several new avenues of research.  One important
application of bitext maps is the construction of translation lexicons
\cite{dagan} and, as discussed, translation lexicons are an important
information source for bitext mapping.  It is likely
that the accuracy of both kinds of algorithms can be improved by
alternating between the two on the same bitext.  There are also plans
to build an automatic bitext locating spider for the World Wide Web,
so that SIMR can be applied to more new language pairs and bitext
genres.

\section*{Acknowledgements} 
SIMR was ported to Spanish/English while I was visiting Sun
MicroSystems Laboratories.  Thanks to Gary Adams, Cookie Callahan, Bob
Kuhns and Philip Resnik for their help with that project.  Thanks also
to Philip Resnik for writing the Spanish tokenizer, and hand-aligning
the Spanish/English training bitexts.  Porting SIMR to Korean/English
would not have been possible without Young-Suk Lee of MIT's Lincoln
Laboratories, who provided the seed translation lexicon, and aligned
all the training and test bitexts.  This paper was much improved by
helpful comments from Mitch Marcus, Adwait Ratnaparkhi, Bonnie Webber
and three anonymous reviewers.  This research was supported by an
equipment grant from Sun MicroSystems and by ARPA Contract
\#N66001-94C-6043.

\end{document}